\documentclass[9pt, twocolumn, twoside]{pnas-new}
\usepackage{comment}
\usepackage{algorithm}
\usepackage{mathtools, nccmath}

\newcommand{\ham}{\mathcal{H}}

\newcommand*\hbz{ \hat{\bf z} }
\newcommand*\hbzt{{ \hat{\bf z}(t_n) }}
\newcommand*\bz{ {\bf z} }
\newcommand*\bzt{ {\bf z}(t_n) }

\newcommand*\dbz{\delta \bz}
\newcommand*\dbzdot{\dot{\delta \bz}}
\newcommand*\dbzdott{\dot{\delta \bz}(t_n)}
\newcommand*\dbzt{ {\delta \bz}(t_n) }
\newcommand*\Bell{\ensuremath{\boldsymbol\ell}} 
 
\newcommand*\Bellt{{\ensuremath{\boldsymbol\ell}}(t_n)}

\newcommand*\Hes{\boldsymbol{\mathcal{D}}}

\newcommand*\FF{{\bf F}}

\templatetype{pnasresearcharticle} 

\title{Local error quantification for Neural Network Differential Equation solvers}

\author[a]{Akshunna S. Dogra}
\affil[a]{John A. Paulson School of Engineering and Applied Sciences, Harvard University, Cambridge, Massachusetts 02138, USA}
\author[b]{William T. Redman}
\affil[b]{Interdepartmental Graduate Program in Dynamical Neuroscience, University of California, Santa Barbara, California 93106, USA}

\leadauthor{Dogra} 

\significancestatement{Neural network (NN) differential equation (DE) solvers are an increasingly attractive scientific tool. They produce robust analytical approximations, are well suited for parallelization, and can sidestep the curse of dimensionality. Unfortunately, precise error quantification in the context of NN DE solvers usually involves setting up some external method for obtaining a ``true'' solution and then comparing the NN approximation with it. Given the substantial cost associated with the use of NN DE solvers, this can be a severe limitation on their utility in practical settings. We propose a new method by which the error associated with their predictions can be identified from within the NN DE solvers themselves, \textit{without} the need for external means. Further, we also demonstrate how this capacity can be leveraged to make such NN DE solvers more efficient.
}

\authorcontributions{ASD was primarily responsible for the conception, mathematics, and overall development of this project. WTR was primarily responsible for conducting and studying the numerical experiments. Both authors contributed to the writing}
\authordeclaration{No competing interests}

\correspondingauthor{\textsuperscript{2}To whom correspondence should be addressed. E-mail: asdpsn@gmail.com}


\begin{abstract}
Neural networks have been identified as powerful tools for the study of complex systems. A noteworthy example is the neural network differential equation (NN DE) solver, which can provide functional approximations to the solutions of a wide variety of differential equations. Such solvers produce robust functional expressions, are well suited for further manipulations on the quantities of interest (for example, taking derivatives), and capable of leveraging the modern advances in parallelization and computing power. However, there is a lack of work on the role precise error quantification can play in their predictions: usually, the focus is on ambiguous and/or global measures of performance like the loss function and/or obtaining global bounds on the errors associated with the predictions. Precise, local error quantification is seldom possible without external means or outright knowledge of the true solution. We address these concerns in the context of dynamical system NN DE solvers, leveraging \textit{learnt} information within the NN DE solvers to develop methods that allow them to be more accurate and efficient, while still pursuing an unsupervised approach that does not rely on external tools or data. We achieve this via methods that can precisely estimate NN DE solver prediction errors point-wise, thus allowing the user the capacity for efficient and targeted error correction. We exemplify the utility of our methods by testing them on a nonlinear and a chaotic system each.
\end{abstract}

\dates{This manuscript was compiled on \today}
\doi{\url{www.pnas.org/cgi/doi/10.1073/pnas.XXXXXXXXXX}}

\begin{document}

\maketitle

\thispagestyle{firststyle}
\ifthenelse{\boolean{shortarticle}}{\ifthenelse{\boolean{singlecolumn}}{\abscontentformatted}{\abscontent}}{}

\section*{\label{sec:level1}Introduction \protect\\ }
Systems described by differential equations (DEs) are ubiquitous. Thus, methods for obtaining their solutions are of wide interest and importance. While numerical methods for DEs have existed for centuries, modern advances in machine/deep learning, cluster/parallel computing, etc, have shown neural networks (NNs) can be powerful options for studying complex DEs \cite{weinan18,Sirig18,Raissi20}. 

NN DE solvers generate closed form, functional approximations to solutions of DEs over domains of interest - providing arbitrarily precise and numerically robust approximations. They have even showcased a capacity to sidestep the curse of dimensionality \cite{grohs18}. Further, these methods are amenable to parallelization in ways many discrete and/or iterative methods are inherently incapable of. 

Landmark works envisioning the utility of NNs in the study of dynamical phenomena \cite{narendra92,lag98,kuschewski93,polycarpou91,narendra89} have leveraged such potential for decades. Recently, those methods were adapted to construct NN solvers for Hamiltonian systems, where they proved to be several orders of magnitude more accurate than symplectic Euler solvers at equivalent temporal discretizations \cite{mar20}.

However, there has been a general lack of work on methods for explicitly and precisely approximating the error associated with NN DE solver predictions. The focus is usually on surrogate markers like the loss functions, which provide imprecise, ambiguous, and/or global measures of performance. This adds a layer of uncertainty to understanding the fitness of the prediction, which can often only be resolved by bench-marking the NN predictions against solutions obtained by established methods, possibly defeating the entire purpose of building NN DE solvers and/or leading to untrustworthy predictions.

Further, computational costs can also limit the utility of NN DE solvers, especially when studying low dimensional systems or systems where domain resolution and further manipulation of quantities of interest is not of importance (taking derivatives, compositions, etc). 

The main insight in this work is that NN DE solvers can be capable of accurate and precise error quantification without external tools: information encoded in their loss functions can allow for the explicit estimation of the error in their predictions.
We also show that this \textit{inherent} error quantification capacity helps in obtaining methods for faster and better NN approximations.

These insights are used for constructing methods that can extend the utility and effectiveness of NN DE solver variants proposed in works like \cite{narendra92,lag98,Raissi20,kuschewski93,polycarpou91,narendra89,mar20,Sirig18,weinan18,anshul20}, when such solvers are constructed for studying smooth dynamical systems. We conjecture our strategies should generalize to all NN DE solvers designed for systems that admit piecewise smooth solutions, but we leave this for later work. 

Let $\FF(\mathbf{z} \equiv [z_1, z_2,..., z_D])\equiv[f_1(\bz),f_2(\bz),...,f_D(\bz)]$ be some smooth operator prescribing the evolution of a $D$ dimensional dynamical system over some time domain:
\begin{equation}
    \label{eq:Dynflow}
    \dot {\bf z}  = \FF({\bf z})
\end{equation}
Further, assume that NN approximations are refined by explicitly aiming to minimize the residual of Eq. \ref{eq:Dynflow}.

We shall derive constraints that accurately and precisely quantify the error associated with such NN predictions and prescribe error correction methods that lead to faster and more accurate approximations than the standard NN optimization methods. We exemplify the utility of our ideas on a nonlinear oscillator and the chaotic Henon-Heiles system. We end with a small discussion on why our ideas should generalize significantly beyond smooth dynamical system NN DE solvers.

\section*{\label{NN_speedup} Neural Network Approaches to Smooth Dynamical Systems\protect\\ }
The central aim of this section is to generalize existing results and showcase strategies to significantly magnify the efficiency of existing dynamical system NN solvers, with minimal modifications to the existing methods.

In \cite{mar20}, the authors presented a rapidly convergent NN that could find accurate functional approximations $\hbz(t)$ for the evolution of phase space parameters $\bz(t)$ of various Hamiltonian systems - chaotic and nonlinear systems included - by simply demanding information about the initial phase space co-ordinate $\bz(0)$ and the temporal domain $[0,T]$ of interest. They demonstrated how NN training can be accelerated by tailoring the architecture for the problem at hand (\textit{physical insight} increasing NN accuracy by advising the choice of activation functions) and how physical parameters of inherent significance can be studied/approximated better by involving the advances that machine/deep learning methods have made in the past few decades (NNs bettering \textit{physical insight} by accurately approximating the dynamics at hand). 

The NN itself was structurally simple and easy to train (Fig. 1): a single unit input layer taking a set $\{t_n\}$ of $M+1$ points from the temporal domain of interest $[0,T]$ each training iteration, two hidden layers with $\sin(\cdot)$ activation units, and an output layer with $D$ outputs ${\bf N} \equiv \{N_1, N_2, ..., N_D\} $ -- one for each state parameter described in Eq. \ref{eq:Dynflow}. Thus, the NN was a $D$ -- dimensional output map for any input $t\in[0,T]$. Sourcing $t_n\in[0,T]$ meant the NN was being trained to give an effective functional approximation $\hbz(t)$ for the expected evolution $\bz(t)$ of the dynamical system over $[0,T]$. To enforce the initial conditions during the training process, the NN output units ${\bf N}(t)$ were related to the ultimate NN prediction $\hbz(t)$ by: $\hbz(t) = \bz(0) + (1-e^{-t}){\bf N}(t)$.

\begin{figure}
    \centering
    \includegraphics{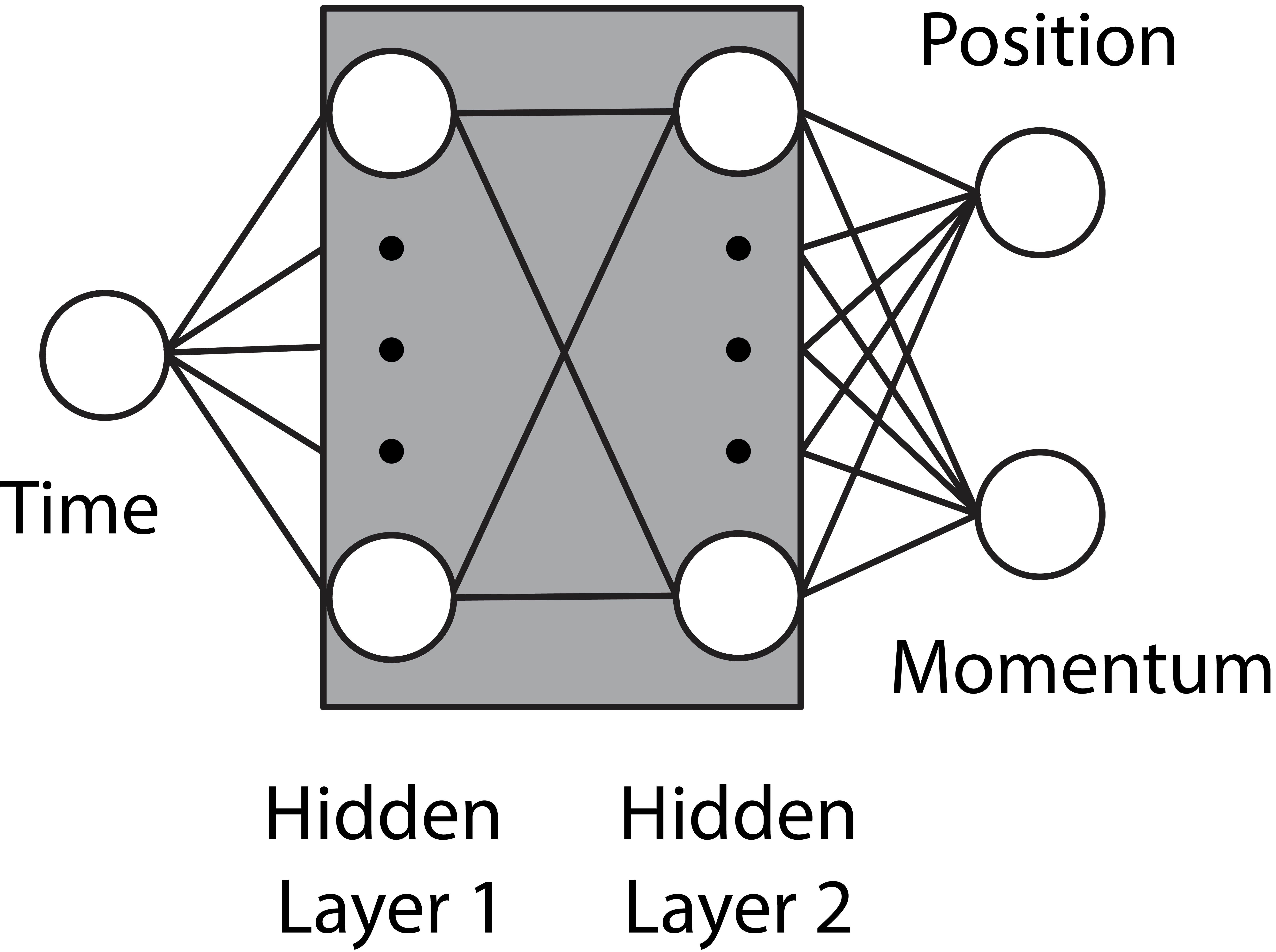}
    \caption{\textbf{Schematic of the Hamiltonian neural network \cite{mar20}}.}
    \label{fig:my_label}
\end{figure}

The authors of \cite{mar20} used the residual of Hamilton's equations as the basis for a mean squared temporal loss
\begin{equation}
    \label{mar20_loss}
    L = \overline{\Bell(t_n) \cdot \Bell(t_n)} \text{ :}\text{ }\text{ }\text{ }\text{ } \Bell (t_n) = {\dot{\hbz}}(t_n) - {\bf J} \cdot (\nabla \ham)|_{\hbzt}
\end{equation}
where $\bf J$ is the symplectic matrix and $(\nabla \ham)|_{\hbzt}$ is the gradient of the Hamiltonian evaluated at the prediction $\hbzt$. Since $\hbz$ is a function of $t$ by the construction of the network, $\dot\hbz$ could be evaluated at any $t_n$ from within the NN, making the training completely unsupervised. 

Eq. \ref{mar20_loss} also implies the capacity of the NN to dispense with causality, since the evaluation of $\Bellt{}$ is not dependent on $\Bell (t_{n-1})$. It also showcases the capacity of NN DE solvers to be parallelized in physical time.

It was shown in \cite{mar20} that $\Bell(t)$ could be re-written as:
\begin{equation}
    \label{eq:Err_anal_1}
    \Bell (t_n) \approx   {\bf J} \cdot \left(\Hes(\ham)|_{\hbz (t_n)}\cdot \delta{\bz}(t_n)\right)   -  \dot{\delta \bz}(t_n)
\end{equation}
where $\Hes(\ham)|_{\hbz (t_n)}$ is the Hessian of the Hamiltonian evaluated at $\hbzt$ and $\delta\bz (t_n)= \bz (t_n) - \hbz (t_n)$ is the difference between the true evolution and the NN prediction. Finally, this reformulation of $\Bell(t)$ was used to obtain explicit global bounds on $\dbz$ over $[0,T]$ (Eq. 24 in \cite{mar20}). 

\subsection*{Local Error Quantification}
Let us describe (and generalize to smooth dynamical systems) strategies for constructing efficient NN DE solvers similar to the ones found in \cite{lag98,mar20}. Let the NN make its predictions for some discrete, finite set of $M+1$ time points $\{t_n \}$, with $t_0=0$ and $t_{M}=T$ at every iteration of training. All intermediate $t_n$ are sampled randomly from $(0,T)$ before each forward pass. We define, following the approach in \cite{lag98,mar20}, the following time averaged cost/loss function $L$
\begin{equation}
    \label{eq:Err_anal_initial}
    L = \overline{\Bell(t_n) \cdot \Bell(t_n)}\text{ :}\text{ }\text{ }\text{ }\text{ }\text{ }\text{ }\text{}\text{ }\text{ }\text{ }\text{ }\text{ } \Bell (t_n) = {\dot{\hbz}}(t_n) - \FF\left(\hbzt\right)
\end{equation}
where $\hbzt$ is the NN prediction and $\FF$ is the dynamical operator of the system. 

The local residual vector $\Bellt$ is the centerpiece of the NN DE solver method: $L \to 0$ implies $\Bellt\to \bf 0$, for all $t_n\in [0,T]$. This implies $\hbzt\to\bzt$, if the initial condition is being enforced, which can be handled by parametrizations like in \cite{mar20}: $\hbzt=\bz(0)+\left(1-e^{-t}\right){\bf N}(t)$.

Let $\bzt$ be the true value at $t_n$ and $\dbzt = \bzt - \hbzt$. Assume the NN is trained sufficiently such that the following Taylor
expansion of $\bf F$ is convergent in $[0,T]$:
\begin{equation}
\begin{aligned}[b]
\label{eq:Taylor_initial}
&\FF(\bz) = \FF(\hbz) + (\FF_{\bz}|_{\hbz}\cdot\dbz) + \frac{(\dbz ^T\cdot\FF _{\bz\bz}|_{\hbz}\cdot\dbz)}{2!} + ..... \implies\\
&\FF(\hbz) = \FF(\bz) - \left[(\FF _{\bz}|_{\hbz}\cdot\dbz) + \frac{(\dbz ^T\cdot\FF _{\bz\bz}|_{\hbz}\cdot\dbz)}{2!} + ...\right]
\end{aligned}
\end{equation}
Here, $\FF_{\bz}\equiv \nabla \FF$, $\FF_{\bz\bz}\equiv \nabla(\nabla \FF)$ and so on, and are being evaluated at $\hbzt$. We note that many common dynamical operators are built from elementary functions with infinite or qualitatively large radii of convergence. 

We know that $\forall t_n, {\dot{\bz}}(t_n) - \FF(\bzt) = 0$. We use Eq. \ref{eq:Err_anal_initial} and \ref{eq:Taylor_initial} to generalize Eq. \ref{eq:Err_anal_1} (Eq. 15 in \cite{mar20}):
\begin{equation}
\begin{aligned}[b]
    \label{eq:Err_anal_reworked}
&\Bell (t_n) = \big[(\FF _{\bz}|_{\hbzt}\cdot\dbzt) + \frac{(\dbz^T(t_n)\cdot\FF _{\bz\bz}|_{\hbzt}\cdot\dbzt)}{2!}
 \\
&\text{ }\text{ }\text{ }\text{ }\text{ }\text{ }\text{ }\text{ }\text{ }\text{ } + ...\big] - \dbzdott
\end{aligned}
\end{equation}
Let us say that $\Bellt\cdot \Bellt \leq l^2_{max}$ and $\sigma _{min}$ is the minimum singular value of $\FF _{\bz}$ for $t_n\in [0,T]$. Keeping only the leading order $\FF _{\bz}|_{\hbzt}\cdot\dbzt$ term in Eq. \ref{eq:Err_anal_reworked}, we get
\[
\Bell (t_n) = \FF _{\bz}|_{\hbzt}\cdot\dbzt - \dbzdott
\]
Let the maxima of $\dbz\cdot\dbz$ be obtained for some $t'\in(0,T)$, i.e., not on the boundary of our time domain of interest. Clearly, $\dbz(t')\cdot\dbzdot(t') = 0$, since $t'$ is not on boundary. This implies 
\[
\Bell (t')\cdot\Bell (t') = (\FF _{\bz}|_{\hbz(t')}\cdot\dbz(t'))^2 + (\dbzdot(t'))^2 \leq l^2_{max}
\]
With $\sigma_{min}$ as minimum singular value of $\FF_{\bz}$, for $t_n\in[0,T]$:
\begin{equation}
\label{generalized_error_bounds}
|\dbz| \leq \frac{{l_{max}}}{\sigma_\text{min}} 
\end{equation}
Thus, we significantly improve and generalize bounds in Eq. 24 of \cite{mar20}, under the assumption that the reduced order form of Eq. \ref{eq:Err_anal_reworked} can offer adequate descriptions of $\dbz$.

Eq. \ref{generalized_error_bounds} gives us a way of using the loss function to bound the error in the predictions of the NN DE solver. However, given smoothness for our operator $\FF$ and the sufficient training assumption used to obtain Eq. \ref{eq:Err_anal_reworked}, we can derive more than just bounds on expected NN error.

Eq. \ref{eq:Err_anal_initial} tells us $\Bell(t)$ is a smooth function (since $\hbz$ is a smooth function of $t$ by the construction of our NN and the operator $\FF$ is assumed to be a smooth operator). Further, the NN can calculate $\Bellt, \FF _{\bz}|_{\hbzt}, \FF _{\bz\bz}|_{\hbzt}, ...$ for any $t_n$. At the end of any training iteration, ${\bf N}(t)$ and $\hbz(t)$ are fixed closed form expressions, which implies $\dbz(t)$ is now a determinable trajectory like $\bz(t)$ and its evolution is governed by Eq. \ref{eq:Err_anal_reworked}. 

Let $\Delta t_n = t_{n+1} - t_n,\text{ } \dbz (0) = 0$. We use the following discrete, recursive relation to estimate $\dbzt$ by picking a small enough $\Delta t_n$, such that $\delta {\bz}(t_n)$ is reasonably resolved (such $\Delta t_n$ exists due to smoothness of $\Bell$ and \FF)

\begin{equation}
\begin{aligned}[b]
    \label{Err_anal_fixing_term}
     &\dbz(t_{n+1}) = \dbzt + \Delta t_n\Big\{\big[(\FF _{\bz}|_{\hbzt}\cdot\dbzt)\\
     &+ \text{ }\text{ }\text{ }\frac{1}{2!}(\dbz^T(t_n)\cdot\FF _{\bz\bz}|_{\hbzt}\cdot\dbzt) + ...] - \Bellt \Big\}
\end{aligned}
\end{equation}
We can use Eq. \ref{Err_anal_fixing_term} to estimate $\dbz$ during any stage of training, to as good a resolution and accuracy as needed, by choosing an adequately small ${\Delta t_n}$. Thus, dynamical system NN DE solvers contain all the information needed to quantify the fitness/error in their predictions. The implementation is encapsulated in Algorithm 1A.

\subsection*{Efficient Error Correction} 
Eq. \ref{eq:Err_anal_reworked} and \ref{Err_anal_fixing_term} allow other powerful possibilities - one is using a second NN to produce a smooth, functional approximation $\delta \hat{\bz}$ for the error trajectory $\dbz(t)$ associated with any prediction $\hbz(t)$. $\delta{\hat{\bz}}(t)+\hbz(t)$ retains all the advantages of being an NN approximation to the solution of the DE, while being significantly more accurate. Further, the latter half of standard training gives diminishing returns in accuracy, thanks to a saturation of an NN's predictive power, machine precision limitations and/or other factors that may be at play. Eventually, computational resources are better spent on adjusting for the error in the NN prediction, rather than trying to minimize it further. 

Eq. \ref{Err_anal_fixing_term} hints directly at an error correction method - using the new NN as a regression tool (Algorithm 1B). Algorithm 1B cuts down the computational costs of calculating a differential term like ${\dot{\hbz}}(t_n)$ every iteration. For NN DE solvers \cite{narendra92,lag98,Raissi20,kuschewski93,polycarpou91,narendra89,mar20,Sirig18,weinan18,anshul20}, appropriate variants of such terms lead to the dominant costs per iteration -- the differential term $\dot{\hbz}$ has to be evaluated from within the NN, before backpropagation (an additional differential cost) can be applied. Hence, Algorithm 1B (which does not include such calculations) is faster than the existing methods.
\begin{algorithm}[t]
\caption{Local error prediction and correction}
\begin{enumerate}
    \item[]  \textbf{1A: Error Prediction:}
    \item Train the NN DE solver for $K$ iterations, until $\hbzt$ is reasonably within the radius of convergence of $\bzt$ for all $t_n$ (Eq. \ref{generalized_error_bounds} can aid in identifying a ``good enough'' $L$).
    \item Generate $\Bellt$ and $\hbzt$ at $kM$ uniformly distanced points in $[0,T]$ for some adequately large $k\geq 2$ (including at $0$ and $T$).
    \item Generate a data set $\dbz_{ec}(t_n)$ using Eq. \ref{Err_anal_fixing_term}, and the values of $\Bellt$ and $\hbzt$ that were obtained from Step 2.
    \item[] \textbf{1B: Error Correction:}
    \item Save the weights and biases of the original NN DE solver obtained after $K$ iterations of training.
    \item Duplicate the original solver. Let its output be $\mathbf{N_2}$. Reinitialize weights and enforce $\delta\hat{\bz}=(1-e^{-t}){\bf N_2}(t)$.
    \item Every training iteration, assemble a batch consisting of $\dbz_{ec}(t_0) = 0$, $\dbz_{ec}(t_M)=\dbz_{ec}(T)$, and $M - 1$ randomly selected $\dbz_{ec}(t_n)$ from the dataset generated in Step 2.  
    \item Define the local loss vector for the duplicate NN: \\ 
    $\Bell_2(t_n) = \delta\bz_{ec}(t_n) - \delta\hbzt$
    \item Train using $L_2=\overline{\Bell_2(t_n)\cdot\Bell_2(t_n)}$. 
    \item Repeat Step 6 - 8 until requisite accuracy is achieved.
    \item Use $\bz(0)+(1-e^{-t})[{\bf N}(t)+{\bf{N_2}}(t)]$ as the solution approximation for the DE at end of training.
\end{enumerate}
\end{algorithm}

An important choice is $k$ (Step 2 of Alg. 1). The setup cost of Alg. 1A is roughly equivalent to $k$ training iterations. However, simple combinatorics dictates that the NN could train using $\Bell_2(t_n){}$ for much more than $k$ iterations, if $k$ and/or $M$ were large enough. In particular, a large enough $k$ reduces likelihood of similar batches, thus mimicking stochastic gradient methods.

To put this into perspective, for the NN presented in \cite{mar20}, $k=2$ would almost ensure that an exactly repeated batch never occurs with random selection. The utility of $k=50$ would practically last forever, before over-fitting concerns start building up in any appreciable sense. The cost, equivalent to $50$ iterations, is insignificant, since NN DE solver variants train over tens of thousands of full training iterations. We use $k=50$ in our examples.

Existing NN DE solvers give diminishing returns on accuracy as soon as they start settling around a local loss minima. Further training leads to better performance at increasingly untenable computational costs. The core idea is to use those resources for precise, local corrections to the original NN predictions. Using the constraints that govern $\dbz(t)$, we can quantify/better the fitness of our approximations \textbf{without} external resources/references. These ideas should extend to other NN DE solvers too.

We end this section by noting that NN DE solvers can be adept at sidestepping the \textit{curse of dimensionality} \cite{Poggio17} (indeed, for the Black-Scholes equation, this has been rigorously proven \cite{grohs18}). NN DE solvers also exhibit the capacity for producing general representations that can easily adapt to changes in the parameters describing the problem \cite{magill18}. They are a rapidly emerging, yet still novel set of powerful tools for studying complex systems. 
Hence, it is imperative that powerful techniques for quantifying and managing their errors are found.

\section*{A nonlinear and a chaotic example}
We exemplify the utility of local error quantification and correction on two Hamiltonian systems. For both cases we consider, $\FF \equiv {\bf J}\cdot\nabla \ham$, where $\ham$ is the Hamiltonian and ${\bf J}$ is the symplectic matrix
\[\bf J = \begin{pmatrix} 
\bf 0 & \bf I\\
\bf -I & \bf 0
\end{pmatrix}\] and $\bf 0$ and $\bf I$ are the null and identity block matrices. 

The first example we consider is the nonlinear oscillator 
\[\bz\equiv\begin{pmatrix} 
x\\
p_x
\end{pmatrix}, \text{ }\text{ }\text{ }\text{ }\ham \equiv \frac{x^2+p_x^2}{2}+\frac{x^4}{4}\]
The second is the chaotic Henon-Heiles system \cite{Hen64} 
\[\bz\equiv\begin{pmatrix} 
x\\
y\\
p_x\\
p_y
\end{pmatrix}, \text{ }\ham \equiv \frac{x^2+y^2+p_x^2+p_y^2}{2}+\frac{y(3x^2-y^2)}{3}\]
The central objectives of this section are to:
\begin{enumerate}
    \item showcase the local error quantification capacity of the NN DE solver from within, using Eq. \ref{Err_anal_fixing_term}.
    \item showcase that the proposed methods are capable of accuracy better than standard methods.
    \item showcase that the proposed methods are more efficient than standard methods.
\end{enumerate}
We test our methods on the NN solver variants described in \cite{lag98,mar20}. For a fair comparison, we do not modify the solver in any way that affects its performance/efficiency. We claim and prove that our methods are capable of quantifying the error in the NN solver approximations from within the setup. Further, we claim that we can leverage this capacity of the NN DE solver to significantly better its efficiency via Alg 1. To verify these claims, we use Scipy's numerical function odeint to generate what can be considered the ground truth/reference solutions for all practical purposes \cite{oliphant07}.

\begin{figure}
   \centering
   \includegraphics[width = 0.5 \textwidth]{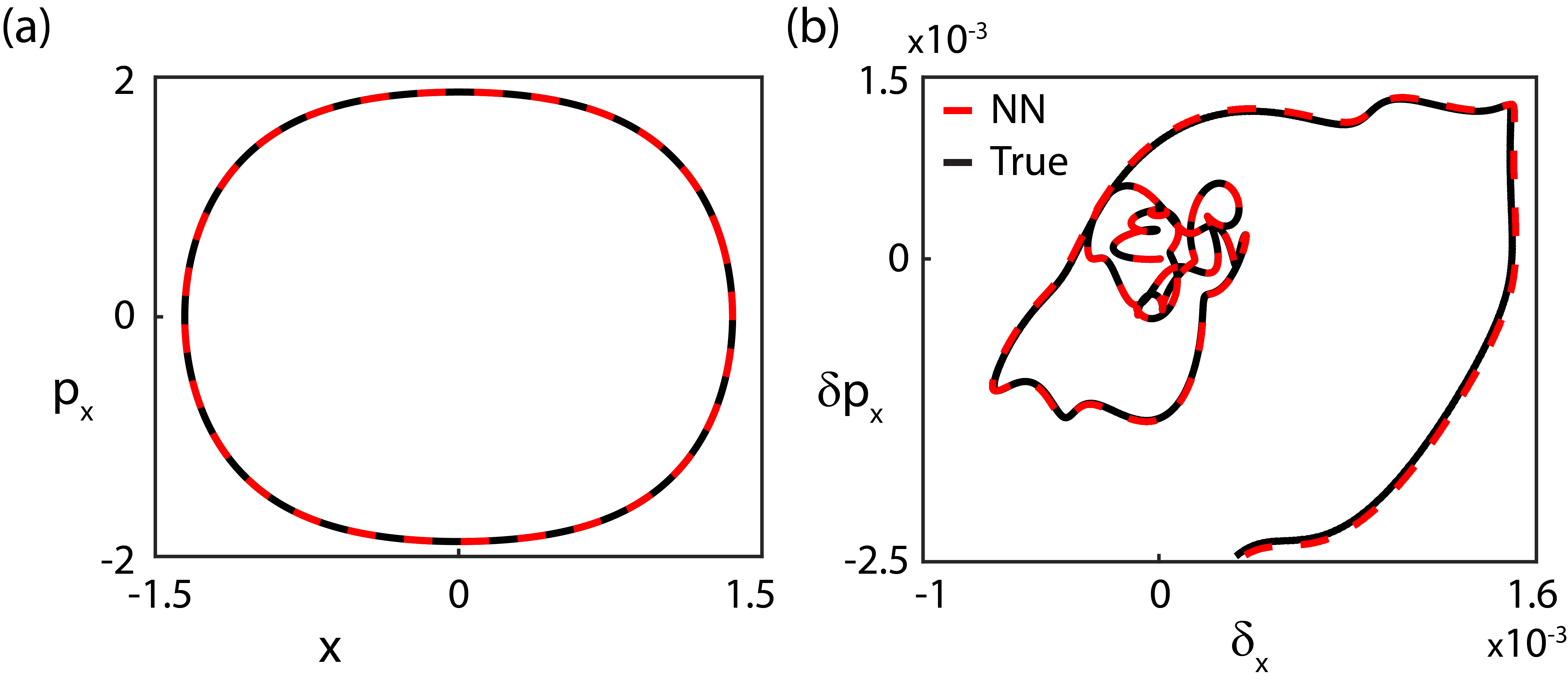}
   \caption{\textbf{Nonlinear oscillator}. (a) True phase space trajectory $\bz$ (black) and NN estimate $\hbz$ (red) using Alg. 1.(b) true NN error trajectory $\delta\bz$ (black) and internal NN estimate $\delta\hat{\bz}$ (red) using Eq. \ref{Err_anal_fixing_term}, after 50,000 iterations of standard training are completed. \textit{Inset}: Zoomed in around $(0, 0)$.
   }
    \label{fig:my_label2}
\end{figure}

\begin{figure*}
   \centering
   \includegraphics[width = 0.7\textwidth]{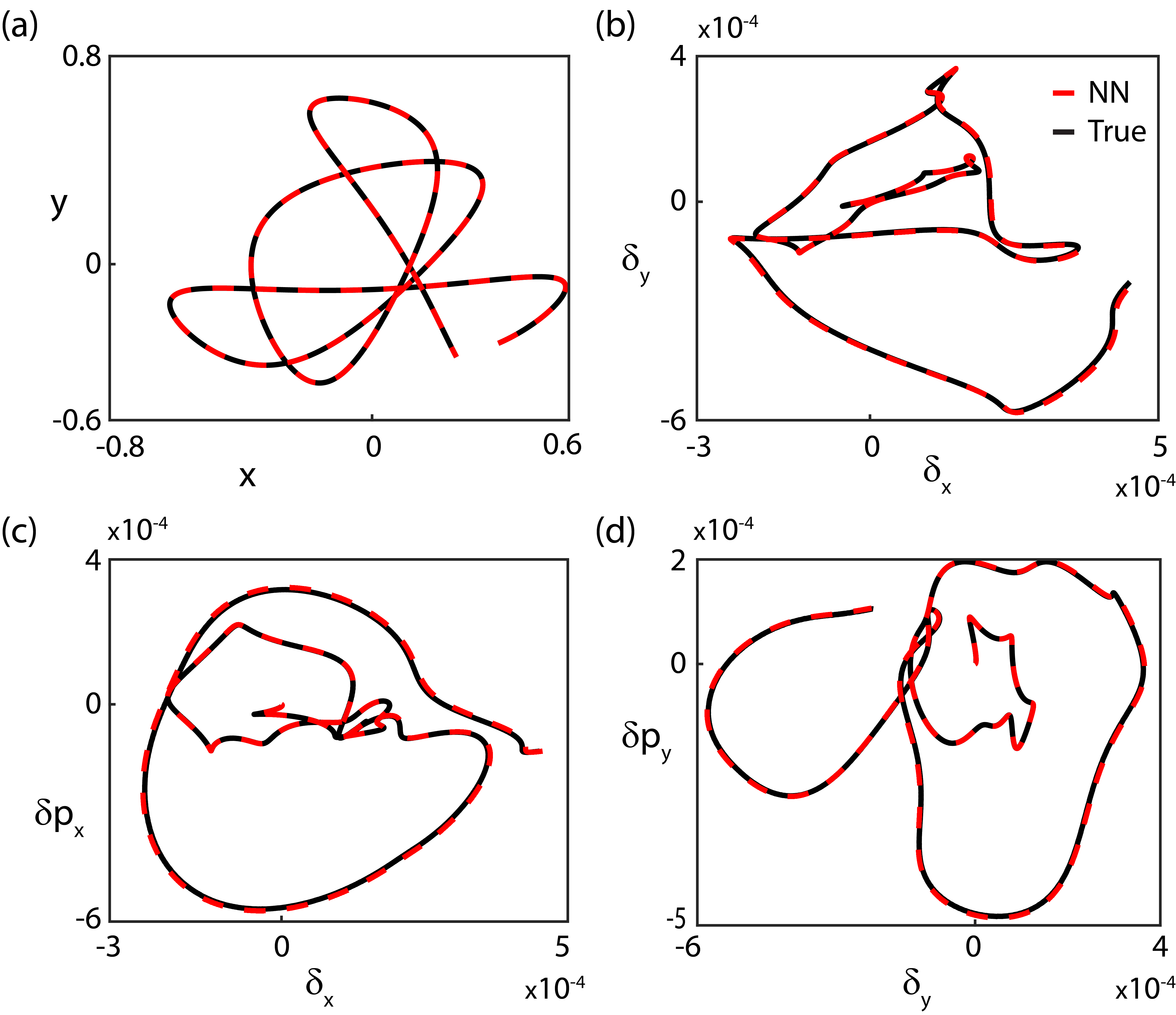}
   \caption{\textbf{Henon-Heiles}. (a) True spatial trajectory $(x,y)$ (black) and NN estimate $(\hat{x},\hat{y})$ (red) using Alg 1. (b,c,d) True NN error trajectories $(\delta x, \delta y), (\delta x,\delta p_x), (\delta y,\delta p_y)$ (black) and internal NN estimates $(\delta \hat{x}, \delta \hat{y}), (\delta \hat{x},\delta \hat{p_x}), (\delta \hat{y},\delta \hat{p_y})$ (red) using Eq. \ref{Err_anal_fixing_term}, after $K = 50,00$0 iterations of standard training are completed.
   }
    \label{fig:my_label3}
\end{figure*}

We setup 100 randomly initialized NNs for each of the two Hamiltonian systems. Phase space initial conditions were randomly chosen as well (see Materials and Methods for details). To begin, we use the solver proposed in \cite{mar20} to train each unique NN solver for 50,000 iterations. We make a very minor modification to this otherwise exact copy of the original setup in \cite{mar20} -- we save the weight/bias values after 20,000 iterations of NN training were completed (this minimal modification adds a one time, insignificant computational cost to the training, with no effect on the accuracy). 

We create copies of each NN solver and initialize them at weight/bias values that were saved at the end of 20,000 iterations of training. These are used to generate the $\dbz_{ec}$ data associated with the predictions of the original NN DE solvers after 20,000 iterations of training were completed. These copies are then kept fixed and the secondary NNs used to fit the $\dbz_{ec}$ data are trained for 30,000 iterations. Thus, we are comparing the effectiveness of our proposed training methods against that of the standard methods, if Algorithm 1 was used with a choice of K = 20,000. Fig 2(a) and 3(a) visually verify the accuracy of our methods. 

We use several measures to compare performance: per iteration run-time $\tau$, mean error $ {\dbz_{avg}}$ and maximum error $\dbz_{max}$. Per iteration run-time $\tau$ is the total cost of setting up and using a method, divided by the number of iterations of use. $ {\dbz_{avg}}$ of the approximation is the mean of $|\dbz|$ over $t\in[0,T]$. $\dbz_{max}$ is the maximum value $|\dbz|$ takes over $t\in[0,T]$. 

${\dbz(t)}$ can be approximated \textbf{internally} from the standard NN DE solver using Eq. \ref{Err_anal_fixing_term} and externally by comparing $\hbz(t)$ with {odeint}'s \textit{ground truth} $\bz(t)$. Fig. 2 (b) and 3 (b, c, d) verify this inherent capacity of the NN.

In Table \ref{tab:my_label}, we report the median observed performance for each method/dynamical system pair. The results are in line with claims made in the previous section: proposed methods outperform existing ones used in \cite{mar20}. Thus, we have verified our claims of NN DE solvers being able to precisely estimate the errors associated with their own predictions. We have also verified that this insight can be leveraged to design better NN approximations.
\begin{table}
    \centering
    \caption{\textbf{Performance comparison for $K = 20,000$.} }
    \begin{tabular}{lrrrr}
         System & Training & Median $\tau$ & Median $ \dbz_{avg}$ & Median $\dbz_{max}$ \\
           &  & ($10^{-3}$ s) & ($10^{-4}$) & ($10^{-3}$) \\
         \toprule
         NL Osc & Standard & 4.98 & 4.58 & 2.03 \\
         NL Osc & {Alg}. \#1 & 2.22 & 2.22 & 0.86 \\
         \midrule 
         HH & Standard & 8.31 &  0.76 & 0.24\\ 
         HH & {Alg}. \#1 & 3.12 & 0.61 & 0.19\\
         \bottomrule
    \end{tabular}
    \label{tab:my_label}
\end{table}

Another regime where local error quantification and correction could be especially useful is when the original NN solver has completely saturated its capacity for accuracy. We designed an experiment to test whether our methods can continue to provide improvements in such a regime. Empirical evidence suggested that the NN DE solver in \cite{mar20} gave very marginal gains in accuracy after 50,000 iterations of training. Hence, we conducted an experiment similar to the previous one, with a choice of $K=50,000$ instead of $20,000$. Both the standard and our methods were then run for an additional $50,000$ iterations, making for a total of 100,000 iterations of training in each case. Table \ref{tab:my_label2} summarizes our results.

\begin{table}
    \centering
    \caption{\textbf{Performance comparison for $K = 50,000$.} }
    \begin{tabular}{lrrrr}
         System & Training & Median $\tau$ & Median $ \dbz_{avg}$ & Median $\dbz_{max}$ \\
           &  & ($10^{-3}$ s) & ($10^{-4}$) & ($10^{-3}$) \\
         \toprule
         NL Osc & Standard & 5.05 & 2.37 & 0.78 \\
         NL Osc & {Alg}. \#1 & 2.23 & 0.77 & 0.24  \\
         \midrule
         HH & Standard & 7.33 & 1.10 & 0.30\\ 
         HH & {Alg}. \#1 & 2.72 & 0.34 & 0.09\\
         \bottomrule 
    \end{tabular}
    \label{tab:my_label2}
\end{table}

It is clear that the NN DE solver in \cite{mar20} is unable to provide major improvements in accuracy after 50,000 training iterations (compare Table \ref{tab:my_label} with Table \ref{tab:my_label2}). In contrast, Table \ref{tab:my_label2} clearly demonstrates that our methods retain and even improve upon their capacity for error correction once the NN DE solver has completely saturated its capacity for the same. As mentioned, this is because the original NN DE solver saturates its capacity for accuracy. Error quantification capabilities allow us to use a secondary NN for the purposes of error correction, by providing fine adjustments to the \textit{coarser} approximation of the primary NN DE solver. 

\section*{Conclusions}
We showcase the power of error quantification on precise, local scales in the context of smooth dynamical system NN solvers. The local information provided by the residual $\Bell$ of the governing DE leads to methods that can accurately quantify the error in the NN prediction over the domain of interest. We were also able to propose an algorithm that directly uses this information to produce more efficient and accurate approximations than those obtainable by standard NN DE methods. 

These methods should extend naturally to DEs that admit piecewise well behaved solutions, since a natural definition for $\Bell$ exists for NN DE solvers designed for such settings. In turn, $\Bell$ encodes relevant information about the fitness of the NN approximation. Further, appropriate variants of Eq. \ref{eq:Err_anal_reworked} would encode this information in a truly local, point-wise sense: they would be valid at any point in the domain where the NN and true solution are sufficiently close. Hence, systems with a finite number of singularities/discontinuities should still be able to leverage our ideas, as long as the boundary conditions are being satisfied and a unique solution to every given set of constraints exists in some well behaved sense.

To our knowledge, this work is the first attempt to utilize information embedded in the components that make up the cost/loss function of NN DE solvers (indeed, any constraint embedding NNs). We also hope to start a conversation on how the errors associated with NN approximations can be explicitly estimated without external tools, even if the function of interest is not known explicitly.
Future work will focus on exploring whether our ideas can generalize to most well behaved systems and their corresponding NN DE solvers. Future work will also involve exploration of other algorithms to take advantage of the inherent capacity for error quantification possessed by NN DE solvers (we present one such example in the appendix).\\

\section*{Materials and Methods}
\subsection*{Implementation}
The code for the Hamiltonian neural network was retrieved from \cite{mar20}. Custom changes were made in order to implement Algo. 1. All computing was done on a Dell work station with 3.4 GHz Intel Core i7-6700 and 32 GB RAM. 

\subsection*{Initial conditions}
To probe the method developed in this paper over a range of state space, we randomly initialized the initial conditions of both the nonlinear and chaotic examples discussed in the main text. 

For the nonlinear oscillator, we picked $x \in [0.3, 2.3]$ and $p_x \in [0, 2]$ at random. For the chaotic Hernon-Heiles system, we picked $x \in [-0.5, 0.5]$ and $y \in [-0.5, \sqrt{3} (1 - |x|)]$ at random. $p_x$ and $p_y$ were sampled from Gaussian distributions with standard deviation $0.1$ and mean $0.25$ and $0.10$ respectively. 

\acknow{ASD was supported as a Research Fellow by the School of Engineering and Applied Sciences at Harvard University during this research. WTR is supported by a Chancellor's Fellowship from UCSB.}

\showacknow

\section{Appendix: Another Error correction Scheme}
Eqn. \ref{eq:Err_anal_reworked} also hints at an error correction method similar to the NN DE solvers themselves - setting up a new NN DE solver that seeks to minimize its residual. $\Bellt, \FF _{\bz}|_{\hbzt}, \FF _{\bz\bz}|_{\hbzt}, ...$ can be generated for any $t_n$ using the original solver and the new solver can be structured as an exact copy of the old one, with a minor modification to the final parametrization: $\delta\hat{\bz}(t) = (1-e^{-t}){\bf N_2}(t)$, where $\bf N_2$ are the hubs of the new output layer. 

New DE and NN solver are structurally and functionally similar to their original NN DE solver counterparts: their per iteration costs should be similar. The method does not change, only the quantity of interest does.

\begin{enumerate}
    \item train the NN DE solver for $K$ iterations, until $\hbzt$ is reasonably within the radius of convergence of $\bzt$ for all $t_n$ (Eqn. \ref{generalized_error_bounds} can aid in identification).
    \item Generate $\Bellt$ and $\hbzt$ at $kM$ uniformly distanced points in $[0,T]$ for some adequately large $k\geq 2$, including at $0$ and $T$.
    \item create a copy of the original solver. Reinitialize weights/biases and enforce $\delta\hat{\bz}=(1-e^{-t}){\bf N_2}(t)$.
    \item define the local loss vector for the second solver: \\ 
    \\
    $\Bell_2(t_n) = \Bell (t_n) - [(\FF _{\bz}|_{\hbzt}\cdot\delta{\hat{\bz}}) + \frac{[\delta{\hat{\bz}}^T(t_n)\cdot\FF _{\bz\bz}|_{\hbzt}\cdot\delta{\hat{\bz}}]}{2!}+\text{ }\text{ }\text{ }\text{ }\text{ }\text{ }\text{ }\text{ }\text{ }\text{ }\text{ }\text{ } .....] + \dot{\delta{\hat{\bz}}}$
    \item every iteration, keep $t_0 = 0, t_M=T$ and randomly select $M-1$ other points from the set of $t_n$ for which $\hbzt$ and $\Bellt$ are available (generating data at $k>2$ reduces likelihood of similar or repeating batches, reducing over-fitting concerns).
    \item train the second NN using $L_2=\overline{\Bell_2(t_n)\cdot\Bell_2(t_n)}$.
    \item use $\bz(0)+(1-e^{-t})[{\bf N}(t)+{\bf{N_2}}(t)]$ as the approximation at end of training.
\end{enumerate}

As mentioned before, NN DE solvers reach a plateau in loss performance after a certain amount of training. This may be a result of various factors, one of which is the saturation of accuracy possible with a given NN architecture, etc. The second NN DE solver is intended for stepping in once a situation like this is reached - by fixing the first NN as a relatively coarser approximation, the second NN seeks to provide smaller scale, finer adjustments that are needed to achieve further accuracy. 

\end{document}